\def\year{2021}\relax
\documentclass[letterpaper]{article} 
\usepackage{aaai21}  
\usepackage{times}  
\usepackage{helvet} 
\usepackage{courier}  
\usepackage[hyphens]{url}  
\usepackage{graphicx} 
\usepackage{mathtools}
\usepackage{commath}
\usepackage{amsmath}
\usepackage[switch]{lineno}
\usepackage{natbib}  
\usepackage{caption} 
\title{Temporal Relational Modeling with Self-Supervision for Action Segmentation}
\author{
    Dong Wang\textsuperscript{\rm 1},
    Di Hu\textsuperscript{\rm 2,3}\thanks{Corresponding author. The research work is partially conducted when the first author was an intern at Baidu Research.},
    Xingjian Li\textsuperscript{\rm 4},
    Dejing Dou\textsuperscript{\rm 4}
    \\
}
\affiliations{

    \textsuperscript{\rm 1}School of Computer Science and Center for OPTical IMagery Analysis and Learning (OPTIMAL),\\ Northwestern Polytechnical University, China\\
    \textsuperscript{\rm 2}Gaoling School of Artificial Intelligence, Renmin University of China, Beijing 100872, China\\
    \textsuperscript{\rm 3}Beijing Key Laboratory of Big Data Management and Analysis Methods\\
    \textsuperscript{\rm 4}Big Data Laboratory, Baidu Research\\
    nwpuwangdong@gmail.com, dihu@ruc.edu.cn, \{lixingjian, doudejing\}@baidu.com
}

\begin{document}

\maketitle

\begin{abstract}
Temporal relational modeling in video is essential for human action understanding, such as action recognition and action segmentation. Although Graph Convolution Networks (GCNs) have shown promising advantages in relation reasoning on many tasks, it is still a challenge to apply graph convolution networks on long video sequences effectively. The main reason is that large number of nodes (i.e., video frames) makes GCNs hard to capture and model temporal relations in videos. To tackle this problem, in this paper, we introduce an effective GCN module, Dilated Temporal Graph Reasoning Module (DTGRM), designed to model temporal relations and dependencies between video frames at various time spans. In particular, we capture and model temporal relations via constructing multi-level dilated temporal graphs where the nodes represent frames from different moments in video. Moreover, to enhance temporal reasoning ability of the proposed model, an auxiliary self-supervised task is proposed to encourage the dilated temporal graph reasoning module to find and correct wrong temporal relations in videos. Our DTGRM model outperforms state-of-the-art action segmentation models on three challenging datasets: 50Salads, Georgia Tech Egocentric Activities (GTEA), and the Breakfast dataset. The code is available at https://github.com/redwang/DTGRM.
\end{abstract}

\section{Introduction}
Action understanding and prediction are fundamental to accomplishing effective communication and interaction between human beings. And the ability to reasoning the temporal relations between actions over time is crucial for human action understanding in daily life. Therefore, temporal relational reasoning in videos is of significant importance for action understanding algorithms, which is the key component in many artificial intelligence systems, such as robot vision \cite{kruger2007meaning,koppula2015anticipating}, intelligent surveillance \cite{danafar2007action}, and autonomous vehicles \cite{rasouli2019autonomous,sadigh2016planning}.

Video-based action segmentation \cite{fathi2011understanding,fathi2013modeling,kuehne2016end,lea2016segmental,lea2017temporal} is the core task for human action understanding, which aims at temporally locating and recognizing human action segments (constituting by consecutive frames with same action labels) in long untrimmed videos, and is much more difficult than action recognition task. The temporal relations between sequential human actions play an important role in action segmentation, because the sequential human actions in daily life always constitute one meaningful event (e.g., making breakfast contains making salad, toasting bread, drinking milk, and etc.).

The topic of action segmentation has been studied by many researchers in the computer vision field. Earlier approaches \cite{rohrbach2012database,karaman2014fast,cheng2014temporal} tried to improve the discriminability of the representations of single frame or video clip and predicted the action label based on learned representations, ignoring the temporal relations between actions. Segmental models \cite{pirsiavash2014parsing,lea2016segmental} and recurrent networks \cite{huang2016connectionist,singh2016multi} paid attention to local temporal dependencies between consecutive actions in videos, and have been demonstrated to have difficulty in modeling long-range temporal relations. Recently, GCNs \cite{huang2020improving} were introduced to improve action segmentation results via modeling temporal relations between pre-computed action segments, while it still focused on the temporal relations among local consecutive action segments. In fact, temporal relations in various timescales (i.e., short-term and long-term timescales) are all of importance to infer action label of each frame. For example, when cooking, people usually first turn on the rice cooker, then cut vegetables and stir fry a few dishes, and at last turn on the rice cooker. There are temporal relations occurring on different timescales, e.g., turn on/off rice cooker, cut different vegetables for one dish, and cut and stir fry actions for one dish. Therefore, capturing and modeling temporal relations in various timescales effectively are at the core of action segmentation and remain difficult for existing methods.

In this work, we propose a Dilated Temporal Graph Reasoning Module (DTGRM) to capture and model the temporal relations and dependencies among actions in different timescales. Further, to enhance temporal reasoning ability of the proposed model, an auxiliary self-supervised task is introduced to identify the wrong-ordered frames in video and predict the correct action labels for them. Specifically, we construct multi-level dilated temporal graphs to effectively capture temporal relations in different timescales, and conduct temporal relational reasoning on the dilated temporal graphs with two complementary edge weights. In the multi-level dilated temporal graphs, we view each video frame as a graph node and update the frame-wise feature representations via the proposed dilated graph reasoning module. Moreover, the auxiliary self-supervision signals are automatically generated by randomly exchanging a fraction of frames in video. By jointly optimizing the auxiliary self-supervised objective function and traditional classification loss function (i.e., cross-entropy loss), the proposed model can effectively learn temporal relations of actions from different time spans, resulting in an improvement on the action segmentation predictions.

The proposed model is evaluated on three challenging benchmark datasets. The experimental results demonstrate the proposed DTGRM is capable of capturing temporal actions and dependencies between video frames in different timescales. Especially, the proposed model outperforms the state-of-the-arts on structure evaluation metrics, i.e., segmental edit score and segmental overlap F1 score. To summarize, the main contributions of this work include:

\begin{itemize}
\item The proposed DTGRM construct multi-level dilated temporal graphs on video frames to effectively model temporal relations in various timescale, and compute two complementary edge weights to conduct temporal relational reasoning with GCNs.
\vspace{-0.1cm}
\item An auxiliary self-supervised task is proposed to enforce the proposed model focus on temporal relational reasoning, which improves the accuracy of the prediction and alleviates the over-fitting problem.
\vspace{-0.1cm}
\item Experiments on multiple benchmark datasets demonstrate the effectiveness of the proposed DTGRM for addressing action segmentation task.
\end{itemize}

\section{Related Work}
\textbf{Action Segmentation}
Action segmentation aims at temporally locating and recognizing multiple action segments in long untrimmed videos. To address this problem, earlier approaches \cite{rohrbach2012database,karaman2014fast} employed the temporal sliding windows to detect the action segments with different lengths. Fathi et al. \cite{fathi2011understanding,fathi2011learning,fathi2013modeling} attempted to use a segmental model to predict the temporally consistent action segments. Cheng et al. \cite{cheng2014temporal} adopted a hierarchical Bayesian non-parametric model to model the temporal dependency between action segments. However, the optimization of these temporal models are mostly time-consuming. 

Other approaches tried to accomplish action segmentation task by predicting action label for every frame in the video. Lea et al. \cite{lea2017temporal} first proposed to use temporal convolution networks (TCN) to predict frame-wise action labels. Lei and Todorovic \cite{lei2018temporal} further proposed deformable temporal convolutions equipped with residual connections to replace the regular temporal convolutions. In addition, Farha et al. \cite{farha2019ms} proposed to use dilated TCN to model the long-range temporal dependencies in videos, and refine the prediction via a multi-stage framework. Recently, Huang et al. \cite{huang2020improving} exploited the temporal relations among multiple action segments with graph convolution networks. However, this method constructed the graph by viewing single action segment from backbone model as one node in graph, which may be very noisy for modeling temporal relations since the prediction from backbone model are mostly inaccurate, resulting in inefficient optimization for GCNs.

\noindent
\textbf{Relational Reasoning with GCNs} 
The graph convolution network (GCN) was proposed by Kipf et al. \cite{kipf2016semi} and has been proved to be effective in modeling the relations in data \cite{li2018beyond,liang2018symbolic}. Recently, GCNs have been widely applied to several research topic in computer vision filed, such as person re-identification \cite{shen2018person}, skeleton-based action recognition \cite{yan2018spatial} and video action recognition and detection \cite{wang2018videos,zhang2020temporal,zhang2019structured,zeng2019graph}. For instance, Zeng et al. \cite{zeng2019graph} proposed to exploit the temporal action proposal-proposal relations using graph convolutional networks. Huang et al. \cite{huang2020improving} improved action segmentation result via modeling temporal relations with GCNs. However, these methods constructed relative small graph based on pre-computed proposals or predicted segments rather than frames. As we all know, the pre-computed proposals and predicted segments are mostly inaccurate and the constructed graphs are noisy. To avoid this problem, in this work, we construct the graphs upon individual frames to achieve more effective relation reasoning.

\noindent
\textbf{Self-Supervision for Video Representation}
The self-supervised pre-trained models and auxiliary self-supervision signals have been proved to be beneficial to many computer vision tasks \cite{doersch2015unsupervised,gidaris2018unsupervised,hu2019deep,hu2020discriminative}. For learning effective video representations with self-supervision, several methods \cite{misra2016shuffle,lee2017unsupervised,fernando2017self} designed auxiliary tasks to verify the input short video clips (i.e., several seconds) is in the correct order or not. These pre-trained models were usually fine-tuned to recognize action on short trimmed videos, while the self-supervised task in this work is specifically designed for action segmentation in long untrimmed videos.

\begin{figure*}
	\begin{center}
		\includegraphics[width=0.95\linewidth]{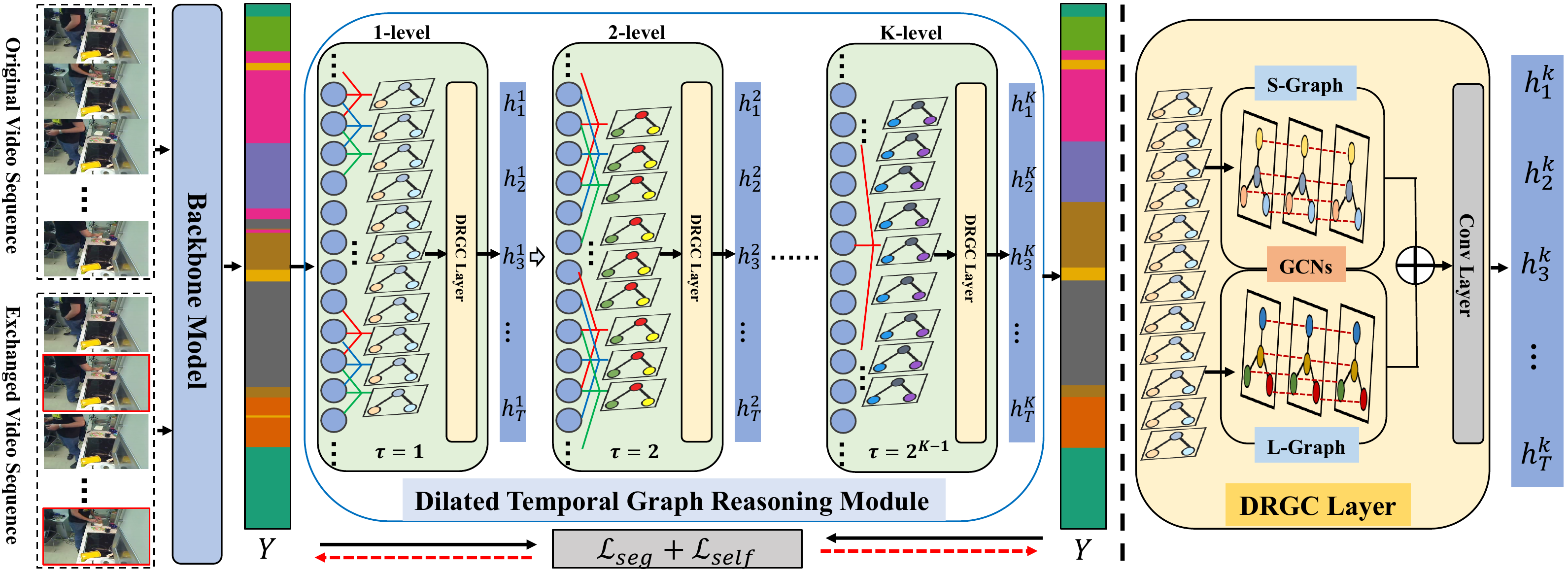}
	\end{center}
	\vspace{-0.2cm}
	\caption{The pipeline of the proposed DTGRM model. The frame-wise features are fed into the backbone model, and the action segmentation results are refined by our DTGRM model. Note that the dilated factor $\tau$ is doubled at each level in DTGRM. $\mathcal{L}_{seg}$ and $\mathcal{L}_{self}$ represent the action segmentation loss and auxiliary self-supervision loss respectively.}
	\label{Fig-pipeline}
\end{figure*}

\section{Our Approach}
We introduce a dilated temporal graph reasoning module (DTGRM) for capturing temporal relations from various timescales in videos, which is essential for the action segmentation task. Given a video of a total $T$ frames, the action segmentation methods need to infer the action class label for each frame $c_{1:T}=(c_1,...,c_T)$, whose ground-truth is given by $y_{1:T}^{gt}=(y_1^{gt},...,y_T^{gt})$, where $y_t^{gt}\in\{0,1\}^C$ is a one-hot vector indicating the true action label. $C$ is the number of action classes including the background class (i.e., no action). Our DTGRM is used to refine the predicted result in an iterative manner, which is built upon a backbone prediction model. In the rest of this section, we first give an overview of the proposed model. Then, the details of our DTGRM and auxiliary self-supervised task are carefully explained.

\subsection{Overview}
The architecture of the proposed model is illustrated in Fig. \ref{Fig-pipeline}. We take the dilated TCN in MS-TCN \cite{farha2019ms} as the backbone model. The backbone model takes frame-wise feature representations $x_{1:T}=(x_1,...,x_T)$, which are extracted using pre-trained I3D network\cite{carreira2017quo}, and outputs the predicted action class likelihoods $y_{1:T}=(y_1,...,y_T)$, where $y_t \in R^C$ are obtained through softmax function. The prediction $y_{1:T}$ is the only input to our DTGRM, which refines the input prediction with GCNs by modeling temporal relations between actions. In addition, inspired by the success on multi-stage refinement \cite{farha2019ms} in action segmentation, we also iteratively refine the prediction using our DTGRM $S$ times to obtain the final prediction result.

In the proposed DTGRM, we view each frame in video as one node and construct multi-level dilated temporal graphs on frames to capture temporal relations in various timescales. Along the constructed multi-level graphs, DTGRM stacks $K$ Dilated Residual Graph Convolution layer (DRGC layer) to conduct temporal relational reasoning on various timescales. Specifically, for each frame in video at each level, we construct two graphs, called S-Graph and L-Graph, on its dilated neighborhood frames. The dilation factor is increasing exponentially while stacking DRGC layers in DTGRM. Note that the edges of dilated graphs represent the relations between frames from various timescales. In the following, a graph with $N$ nodes in GCNs are denoted as $\mathcal{G}(\mathcal{V}, \mathcal{E})$, where $\mathcal{V}$ is the set of the node $v_i$ and $e(i,j)\in \mathcal{E}$ represents the edge weight between node $v_i$ and $v_j$.

Moreover, the over-segmentation problem \cite{lea2016learning} is one of the key factors affecting action segmentation accuracy. To reduce over-segmentation errors in action segmentation results, we introduce an auxiliary self-supervised task to simulate the over-segmentation errors manually. In detail, we first random choose some frames from videos and pairwise exchange them. The goal of the self-supervised task is to identify the exchanged frame and predict the correct action label by temporal relational reasoning in various timescales.

\subsection{Dilated Temporal Graph Reasoning Module}
GCNs have shown promising ability on relational reasoning \cite{chen2019graph,hussein2019videograph,zeng2019graph}. The key step in GCNs is to construct the graphs and compute the edge weights. Previous works usually construct graphs based on action proposals or action segments, which are pre-computed by other models and mostly inaccurate. In contrast, we directly construct graphs on frames and address large graph problem with the proposed multi-level dilated temporal graphs.

\subsubsection{Multi-Level Dilated Temporal Graphs}
Temporal relations from various time spans are very useful to infer action label on single frame, i.e., successive frames always belong to the same action class and long-range temporal relations always capture the relationship between different action classes. But, it is hard to train and optimize GCNs with one large graph containing all frames (i.e., nodes) in videos. To address this problem, we propose to construct multi-level dilated temporal graphs to capture temporal relations between all the frames in videos.

Suppose we have a total of $T$ frames in video and the dilated temporal graphs at $k$-th level are constructed based on dilation factor $\tau_k$. To be specific, for the frame at timestep $t$, its dilated neighborhood frames is $\{ t-\tau_k, t+\tau_k\}$. Then, the frames at time $\{t-\tau_k,t,t+\tau_k\}$ are viewed as nodes and the dilated temporal graph $\mathcal{G}_t^k$ is constructed upon them. We denote the order of the graph (its number of vertices) as $O_t^k$, i.e., $O_t^k=3$. As shown in Fig. \ref{Fig-pipeline}, to capture temporal relations in various time spans, we construct $K$ levels of dilated temporal graphs and apply the proposed DRGC layer at each level, where the dilation factor $\tau_k$ is doubled at each level, i.e.,$\tau_k=2^{k-1}, k\in \{1,2,..,K\}$. Note that all the dilated temporal graphs contain three nodes (i.e., node $v_{t-\tau_k}, v_{t}, v_{t+\tau_k}$). At $k$-th level, to alleviate the noise problem in single constructed graph, we compute two complementary edge weights for dilated temporal graph $\mathcal{G}_t^{(k)}$ and name them as S-Graph $\mathcal{G}_t^{s,(k)}$ and L-Graph $\mathcal{G}_t^{l,(k)}$.

\subsubsection{S-Graph}
The motivation of constructing S-Graph (Similarity Graph) $\mathcal{G}_t^{s,(k)}$ is that the nodes with similar action class likelihoods $y$ should have larger edge weights. Therefore, we first apply one $1\times1$ convolution layer to transfer action class likelihoods $y \in R^C$ into $d$-dimensional hidden representations $h_{1:T}=(h_1,...,h_T)$. Then, for S-Graph $\mathcal{G}^{s,(k)}_t$, the edge weight $e_s(i,j)$ between node $v_i$ and $v_j$ are defined by the cosine similarity between their hidden representations $h_i, h_j$, i.e.,
\begin{equation}
\label{Eq-cosine-similarity}
    e_s(i,j) = \frac{h_i \cdot h_j}{max(\norm{h_i}_2 \cdot \norm{h_j}_2, \epsilon)},
\end{equation}
where $\epsilon$ is a small constant avoiding divide-by-zero. We gather all edge weights in $\mathcal{G}^{s,(k)}_t$ to an adjacency matrix $A^{s,(k)}_t$. The graph convolution operation is used to update the hidden representation $h_t$ of each frame according to its S-Graph $\mathcal{G}^{s,(k)}_t$ at each DRGC layer.

\subsubsection{L-Graph}
Since there mostly are some wrong predictions in action class likelihood $y$ that make the edge weight $e_s(i,j)$ inaccurate, we propose to construct L-Graph (Learned Graph) $\mathcal{G}_t^{l,(k)}$ whose edge weights are generated by one sub-network, which can capture the important temporal relations that are complementary to S-Graph after training. Specifically, we apply one dilated 1D convolution on hidden representations $h_{1:T}$, and the dilation factor of this 1D convolution layer equals to corresponding dilated temporal graph, i.e., $dilation = \tau_k$. The outputs of this layer is the adjacency matrix of graph $\mathcal{G}_t^{l,(k)}$, where the value with index $i,j$ represent the edge weight between node $v_i$ and $v_j$. Formally, the adjacency matrix $A^{l,(k)}_t$ of the graph $\mathcal{G}^{l,(k)}_t$ are defined as
\begin{equation}
\label{Eq-learned-edge}
    A^{l,(k)}_t = Conv(h[t-\tau_k,t,t+\tau_k], W, dilation=\tau_k),
\end{equation}
where $W \in R^{ks \times (O_t\times O_t) \times d}$ are the weights of the dilated convolution filter with kernel size $ks=3$. $O_t=3$ is the number of vertices of the graph $\mathcal{G}^{l,(k)}_t$. The output $A_t^{l,(k)}$ is a $O_t\times O_t$-dimensional vector and reshaped to the adjacency matrix with size $(O_t, O_t)$. Note that the adjacency matrix $A^{l,(k)}_t$ is asymmetric.

\subsubsection{Reasoning on Dilated Temporal Graph}
Given the constructed dilated temporal graphs at each frame $t$ and level $k$, $\mathcal{G}_t^{s,(k)}$ and $\mathcal{G}_t^{l,(k)}$, we apply the proposed DRGC layer on them to conduct temporal relational reasoning in various timescales. To be specific, we first normalize the adjacency matrixes $A_t^{s,(k)}$ and $A^{l,(k)}_t$ with softmax function. Then, for relational reasoning on constructed graphs, our DRGC layers employ the graph convolution layer proposed in \cite{kipf2016semi}:
\begin{equation}
\label{Eq-graph-convolution}
    X = \sigma(AXW),
\end{equation}
where $A \in R^{N\times N}$ is the adjacency matrix of the graph, $X \in R^{N\times d}$ are the hidden representation of all nodes in the graph, and $W \in R^{d\times d}$ is the parameter matrix to be learned. $\sigma$ is the ReLU activation function.

Based on the graph convolution layer and constructed dilated temporal graphs, our DTGRM stacks $K$-level DRGC layers to model temporal relations in various timescales. Specifically, at $k$-th DRGC layer ($k \in \{0,1,...,K-1\}$), the dilated temporal graphs are constructed with dilation factor $\tau = 2^k$. As illustrated in Fig. \ref{Fig-pipeline}, at $t$-th frame, we first separately apply graph convolution on $\mathcal{G}^{s,(k)}_t$ and $\mathcal{G}^{l,(k)}_t$, and then fuse their output with addition operation, i.e.,
\begin{equation}
\label{Eq-DTGRM-layer}
\begin{gathered}
    X_t = h^{(k)}_{[t-\tau_k,t,t+\tau_k]}, \\
    O_t^{(k)} = GCN^{(k)}(X_t, A^{s,(k)}_t, W^{s,(k)}_t) + \\ 
    GCN^{(k)}(X_t, A^{l,(k)}_t, W^{l,(k)}_t), \\
    O_t^{(k)} = o^{(k)}_{[t-\tau_k,t,t+\tau_k]}, \\
    h^{(k+1)}_t = Conv(o_t^{(k)}, W_{(k)}) + h^{(k)}_t,
\end{gathered}
\end{equation}
where $GCN$ is the graph convolution operation defined in Eq. \ref{Eq-graph-convolution}. $A^{s,(k)}_t, A^{l,(k)}_t \in R^{N\times N}$ are the adjacency matrix of the graph $\mathcal{G}^{s,(k)}_t$ and $\mathcal{G}^{l,(k)}_t$. $W^{s,(k)}_t, W^{l,(k)}_t \in R^{d\times d}$ are the parameter matrix of the graph convolution layer for $t$-th frame at $k$-th layer. $W_{(k)} \in R^{1\times d\times d}$ is the weights of the 1D convolution filter with kernel size 1, which is shared with each timestep in video. With stacking the DRGC layer $K$ times, our DTGRMs can efficiently capture short and long-range temporal relations in videos and avoid the large graph problem. In this way, our DTGRMs conduct temporal realtional reasoning in various timescales, which is essential for action segmentation.

To get the action class likelihoods $y_{1:T}$ for each frame, we apply a fully-connected layer over the outputs of the last DRGC layer followed by a softmax activation, i.e.,
\begin{equation}
\label{Eq-classification}
    y_{1:T} = softmax(Wh^{(K)}_{1:T}+b),
\end{equation}
Where $W\in R^{C\times d}$ and $b\in R^C$ are the weights and bias for the FC layer. $h^{(K)}_{1:T}$ is the output of the $K$-th DRGC layer.

\subsection{Auxiliary Self-Supervision}
Self-supervision signals have been used for video representation learning \cite{misra2016shuffle,lee2017unsupervised,fernando2017self,korbar2018cooperative} and improved the downstream tasks, such as action recognition and action detection. Compared to supervised learning methods, self-supervised methods automatically generate the supervisory signals (i.e., pseudo label) that are inferred from the structure of the data, without involving any human annotation. In this work, different from previous works that only provide the self-supervision signals on video-level, we obtain the frame-wise self-supervision signals in the context of the pairwise exchanging frames in video, which simulates the over-segmentation errors in the action segmentation results.

Specifically, given the input video sequence $x_{1:T}=(x_1,...,x_T)$ with correct temporal order. We select $\eta$\% frames and randomly form them as frame pairs $\{x_{t_i}, x_{t_j}\}$, then the frames in each pair are exchanged. The resulting video sequence $x_{1:T}^{ex}=(...,x_{t_j},...,x_{t_i},...)$ contains some wrong ordered frame and is fed into the proposed model along with original video sequence $x_{1:T}$. The outputs corresponding to $x_{1:T}^{ex}$ consist of action class likelihoods $y^{ex}_{1:T}$ and exchange likelihood $e^{ex}_{1:T}$, which are obtained by feeding the hidden representation $h_{1:T}^{K,ex}$ into a fully-connected layer. The goal of the auxiliary self-supervised task is to identify the exchanged frames and predict the correct action labels that should be at their moments. Formally, we generate a binary self-supervised signal $p_{1:T}=(p_1,...,p_T)$ to label the exchanged frames, where $p_t \in \{0,1\}^2$ is the one-hot vector indicating whether $t$-th frame is exchanged or not. Moreover, exchanged frames are the prefect simulation of over-segmentation errors in action segmentation task. Therefore, except the binary training label $p_{1:T}$, we directly take the ordered ground-truth action label $y_{1:T}^{gt}$ as another training label. The overall loss function of self-supervision is
\begin{equation}
\label{Eq-self-sup-loss}
    \mathcal{L}_{self} = \mathcal{L}_{ex}(e^{ex}, p) + \mathcal{L}_{corr}(y^{ex}, y^{gt}),
\end{equation}
where $e^{ex} \in R^{T\times 2}$ and $y^{ex} \in R^{T\times C}$ (for simplicity, we drop the timestep notation). With the above self-supervised objective function, our DTGRM learns to do accurate temporal relational reasoning about the temporal relation structure, leading to better action segmentation results.

\subsection{Training and Loss Function}
We train the backbone model and our DTGRM in an end-to-end manner with a combination of the multiple loss functions. The inputs of the whole network is the ordered video sequence $x_{1:T}$ and exchanged video sequence $x_{1:T}^{ex}$, and the outputs is the action class likelihood $y_{1:T}$, $y_{1:T}^{ex}$ and exchange likelihood $e_{1:T}^{ex}$. As for the action class likelihood $y_{1:T}$ and $y_{1:T}^{ex}$, we apply the typical cross entropy loss
\begin{equation}
\label{Eq-cross-entropy}
    \mathcal{L}_{cls}(y, y^{gt}) = \frac{1}{T}\sum_{t}^{T}\sum_{c}^{C} -y^{gt}_{t,c}log(y_{t,c}).
\end{equation}
And we adopt the truncated mean squared error $\mathcal{L}_{t-mse}$ proposed in \cite{farha2019ms} to punish local inconsistency in action class likelihood. Based on these loss functions, the action segmentation loss for ordered video sequence and auxiliary self-supervised task loss function are defined as follows,
\begin{equation}
\label{Eq-all-loss}
\begin{gathered}
    \mathcal{L}_{seg} = \mathcal{L}_{cls}(y, y^{gt}) + \omega\mathcal{L}_{t-mse}, \\
    \mathcal{L}_{ex}(e^{ex}, p) = \lambda_e\mathcal{L}_{cls}(e^{ex}, p), \\
    \mathcal{L}_{corr}(y^{ex}, y^{gt}) = \lambda_c\mathcal{L}_{cls}(y^{ex}, y^{gt}) + \omega\mathcal{L}_{t-mse}, \\
    \mathcal{L} = \mathcal{L}_{seg} + \mathcal{L}_{self},
\end{gathered}
\end{equation}
where $\omega, \lambda_c, \lambda_e$ are hyper-parameters that balance the components in loss function. Since we apply our DTGRM $S$ times sequentially, the loss function $L$ is applied on the outputs from the each DTGRM and backbone model. 

\section{Experiments}
\subsubsection{Implementation Details}
The whole model proposed in this paper consists of one backbone network and three DTGRMs (i.e., $S=3$) that are implemented with Pytorch library on Nvidia 2080Ti GPU. We set the dimension of hidden representation $d$ as 64 for backbone network and our DTGRMs. The proposed DTGRM constructs $K=10$ dilated temporal graphs and apply DRGC layer on each level, where the dilation factor is doubled at each level. For hyper-parameter $\eta$ in auxiliary self-supervised task, we set it as $\eta=20$. For the loss function, we set $\omega=0.15$, $\lambda_e=2$ and $\lambda_c=0.5$. In all experiments, the network is trained using Adam optimizer with a learning rate of 5e-4.

\subsubsection{Datasets}
The 50Salads \cite{stein2013combining} dataset consists of 50 videos of 17 action classes, which averagely contains 20 action instances and is 6.4 minutes long. The videos capture the salad preparation activities performed by 25 actors where each actor prepares two different salads. The GTEA \cite{fathi2011learning} dataset contains 28 videos with 7 different activities performed by 4 subjects, such as preparing coffee and cheese sandwich. Each video is annotated with 11 fine-grained action classes and averagely has 20 action instances. The Breakfast \cite{kuehne2014language} dataset is the largest among the three datasets with 1712 videos, recording the breakfast related activities in 18 different kitchens. The videos are annotated with 48 different actions and contain 6 action instances on average. In all datasets, we sample the videos with fixed fps rather than fixed number of frames and extract I3D \cite{carreira2017quo} features for the video frames, which are input to the proposed model.

\subsubsection{Evaluation Metrics}
For evaluating our model, we adopt the following evaluation metrics as in \cite{lea2017temporal,farha2019ms,huang2020improving}: frame-wise accuracy (Acc), segmental edit distance (Edit) and segmental F1 score at overlapping thresholds 10\%, 25\% and 50\%, denoted by F1@\{10,25,50\}. The overlapping ratio is the intersection over union (IoU) ratio between predicted and ground-truth action segments. Frame-wise accuracy is the most commonly used metric for action segmentation. However, actions with long duration tend to have a higher impact than actions with short duration on this metric, and there is no explicit penalty on over-segmentation errors. In contrast, segmental edit score and F1 score presented in \cite{lea2017temporal,lea2016segmental} are used to penalizes the over-segmentation errors and measure the quality of the prediction.

\begin{table}[htbp]
	\centering
	\caption{Comparisons with the state-of-the-art methods on 50Salads, GTEA, and the Breakfast dataset.}
	\vspace{-0.2cm}
	\begin{tabular}{cccccc}
		\hline
		\textbf{50Salads} &\multicolumn{3}{c}{\textbf{F1}$@\{ \text{10,25,}50 \}$} &\textbf{Edit} &\textbf{Acc} \\  
		\hline
		MSTCN &76.3 &74.0 &64.5 &67.9 &80.7 \\
		MSTCN++ &80.7 &78.5 &70.1 &74.3 &83.7 \\
		BCN &82.3 &81.3 &74.0 &74.3 &84.4 \\
		MSTCN+GTRM &75.4 &72.8 &63.9 &67.5 &82.6 \\
		\hline
		DTGRM &79.1 &75.9 &66.1 &72.0 &80.0 \\
		\hline
		\hline
		\textbf{GTEA} &\multicolumn{3}{c}{\textbf{F1}$@\{ \text{10,25,}50 \}$} &\textbf{Edit} &\textbf{Acc} \\ 
		\hline
		MSTCN &85.8 &83.4 &69.8 &79.0 &76.3 \\
		MSTCN++ &88.8 &85.7 &76.0 &83.5 &80.1 \\
		BCN &88.5 &87.1 &77.3 &84.4 &79.8 \\
		MTDA &82.0 &80.1 &72.5 &75.2 &83.2 \\
		\hline
		DTGRM &87.8 &86.6 &72.9 &83.0 &77.6\\
		\hline
		\hline
		\textbf{Breakfast} &\multicolumn{3}{c}{\textbf{F1}$@\{ \text{10,25,}50 \}$} &\textbf{Edit} &\textbf{Acc} \\ 
		\hline
		MSTCN &52.6 &48.1 &37.9 &61.7 &66.3 \\
		MSTCN++ &64.1 &58.6 &45.9 &65.6 &67.6 \\
		BCN &68.7 &65.5 &55.0 &66.2 &70.4 \\
		MSTCN+GTRM &57.5 &54.0 &43.3 &58.7 &65.0 \\
		\hline
		DTGRM	&68.7 &61.9 &46.6 &68.9 &68.3\\
		\hline
	\end{tabular}\label{Table-compare-sota}
\end{table}

\subsection{Comparison with the State-of-the-Art}
In this section, we compare the proposed model with several state-of-the-art models on three datasets: 50Salads, GTEA, and the Breakfast dataset. The results are presented in Table. 1. Specifically, the comparison methods consists of five closely related state-of-the-art models, including MSTCN \cite{farha2019ms}, MSTCN++\cite{li2020ms}, MTDA\cite{chen2020action},  BCN \cite{wangboundary}, and MSTCN+GTRM \cite{huang2020improving}. MSTCN are the recent temporal convolution based model that adopted the similar multi-stage framework as our approach (i.e., iteratively refining the prediction from backbone model several times), which is the baseline method of the proposed model. We use the same backbone model and the same num of layers and stages with the MSTCN. MSTCN++ and MTDA are the extended works of MSTCN. BCN improves the smoothness of frame-wise predictions by cooperating action boundary information. MSTCN+GTRM is the most related to our models, where the GCNs are used to model relations between action segments upon the results of MSTCN model.

As can be seen in Table \ref{Table-compare-sota}, the proposed DTGRM model outperforms the baseline method MSTCN on the three datasets and by a large margin with respect to three evaluation metrics. Specifically, our DTGRM model achieves a moderate improvement on 50Salads and GTEA dataset, i.e., around 2-5\% in all evaluation metrics, except the frame-wise accuracy on the 50salads. As for the Breakfast dataset, our approach outperforms MSTCN and MSTCN+GTRM with a larger margin, i.e., near 10\% improvement on F1 score and segmental edit score. This shows that our DTGRM is capable of reducing over-segmentation errors in prediction. In addition, the improvements over MSTCN demonstrate that dilated temporal convolution in MSTCN is inefficient in temporal relations reasoning.

\begin{table}[htbp]
	\centering
	\caption{Comparisons of performance by our DTGRM and its variants on the GTEA, 50 Salads and Breakfast dataset.}
	\vspace{-0.2cm}
	\begin{tabular}{cccccc}
		\hline
		\textbf{GTEA} &\multicolumn{3}{c}{\textbf{F1}$@\{ \text{10,25,}50 \}$} &\textbf{Edit} &\textbf{Acc} \\ 
		\hline
		MSTCN &85.8 &83.4 &69.8 &79.0 &76.3 \\
		DTCN(w/o self) &86.3 &83.6 &70.6 &80.8 &76.1 \\
		S-Graph(w/o self) &48.2 &44.9 &37.4 &38.4 &71.3 \\
		L-Graph(w/o self) &85.6 &83.7 &70.3 &78.8 &76.4 \\
		\hline
		DTGRM(w/o self) &\textbf{87.3} &\textbf{85.5} &\textbf{72.3} &\textbf{80.7} &\textbf{77.5} \\
		\hline
		\hline
		\textbf{50Salads} &\multicolumn{3}{c}{\textbf{F1}$@\{ \text{10,25,}50 \}$} &\textbf{Edit} &\textbf{Acc} \\ 
		\hline
		BK &52.7 &47.8 &40.0 &42.0 &78.0 \\
		BK+1-DTGRM &69.0 &65.3 &56.2 &60.3 &79.3 \\
		BK+2-DTGRM &75.2 &71.6 &62.6 &66.6 &79.7 \\
		BK+3-DTGRM &\textbf{79.1} &\textbf{75.9} &\textbf{66.1} &\textbf{72.0} &\textbf{80.0} \\
		\hline
		\hline
		\textbf{Breakfast} &\multicolumn{3}{c}{\textbf{F1}$@\{ \text{10,25,}50 \}$} &\textbf{Edit} &\textbf{Acc} \\ 
		\hline
		MSTCN(IDT) &58.2 &52.9 &40.8 &61.4 &65.1 \\
		S-Graph(w/ self) &49.6 &43.7 &31.9 &54.6 &66.3 \\
		L-Graph(w/ self) &67.5 &60.7 &45.3 &68.2 &68.0 \\
		\hline
        DTGRM &\textbf{68.7} &\textbf{61.9} &\textbf{46.6} &\textbf{68.9} &\textbf{68.3} \\
        \hline
	\end{tabular}\label{Table-DTGRM}
\end{table}

\begin{figure*}
	\begin{center}
		\includegraphics[width=0.95\linewidth]{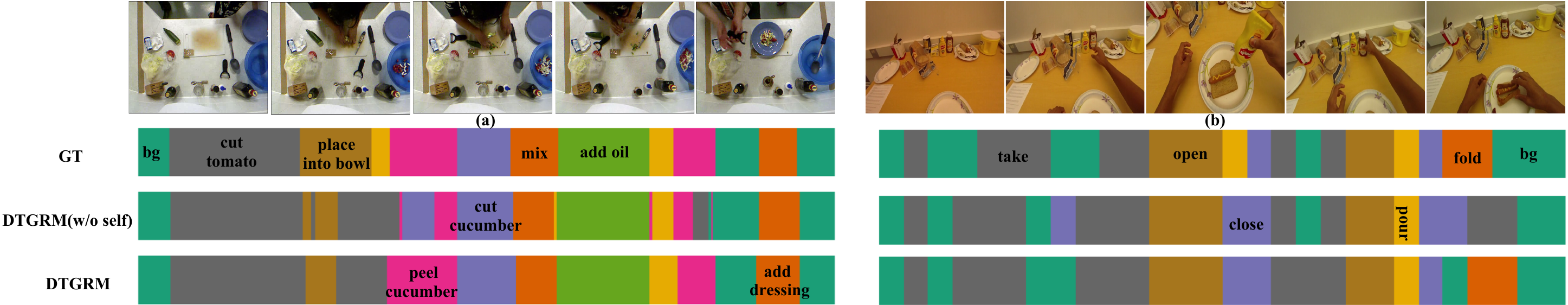}
	\end{center}
	\vspace{-0.2cm}
	\caption{Qualitative comparison of the action segmentation results on (a)50Salads, and (b)GTEA dataset. Only few frames of the whole video are shown for clarity. We can see that the model trained with self-supervision generate better results.}
	\label{Fig-compare-self}
\end{figure*}

\subsection{Ablation Studies}
\subsubsection{The Effectiveness of DTGRM model}
To verify the effect of each constructed graph in our DTGRM, we conduct ablation studies by changing or deleting part of DRGC layer in our DTGRM. All these models are implemented based on the same backbone model and trained without auxiliary self-supervision on GTEA dataset. As shown in upper part of Table \ref{Table-DTGRM}, \textbf{``DTCN''} is the case where GCNs in DRGC layers are replaced by the dilated temporal convolution layer presented in \cite{farha2019ms}. Our DTGRM outperforms this approach by 1-3\% in all metrics, which validates that our method can effectively capture temporal relations from various time spans to improve action segmentation. \textbf{``S-Graph''} indicates the model that only applies GCNs on S-Graph while ignoring the L-Graph. The results of this model suggest that the S-Graph may be very noisy due to the errors in prediction from backbone model. \textbf{``L-Graph''} represents the model that only applies GCNs on L-Graph while ignoring the S-Graph and achieves comparable performance, which shows the learned graph weights are more appropriate and useful to capture the temporal relations in videos. In addition, we compare the results from backbone model (denoted as ``BK'') and models with different number of DTGRMs (denoted as ``BK+*-DTGRM'') that are trained with auxiliary self-supervision on 50Salads dataset. As shown in middle part of Table \ref{Table-DTGRM}, the performance is significantly improved after using only one DTGRM and stacking more DTGRMs can improve the predictions performance on segmental edit distance and segmental F1 score progressively, which demonstrates the effectiveness of our DTGRM on improving the quality of the predictions. The results on Breakfast dataset also prove the effectiveness of the proposed method.   

\begin{table}[htbp]
	\centering
	\caption{Comparisons of performance by our DTGRM and its variants with or without auxiliary self-supervision on the 50Salads dataset.}
	\vspace{-0.2cm}
	\begin{tabular}{cccccc}
		\hline
		\textbf{50Salads} &\multicolumn{3}{c}{\textbf{F1}$@\{ \text{10,25,}50 \}$} &\textbf{Edit} &\textbf{Acc} \\ 
		\hline
		DTCN(w/o self) &74.7 &71.8 &63.9 &66.7 &80.3 \\
		DTCN(w/ self) &79.0 &76.2 &66.4 &71.4 &78.1 \\
		\hline
		S-Graph(w/o self) &52.6 &48.1 &39.9 &41.9 &77.5 \\
		S-Graph(w/ self) &59.3 &54.7 &46.5 &49.5 &78.9 \\
		\hline
		L-Graph(w/o self) &73.5 &71.4 &60.8 &65.3 &77.3 \\
		L-Graph(w/ self) &78.6 &75.6 &66.4 &70.9 &79.5 \\
		\hline
		DTGRM(w/o self) &74.0 &71.0 &60.8 &67.9 &77.9\\
		DTGRM(w/ self) &79.1 &75.9 &66.1 &72.0 &80.0\\
		\hline
	\end{tabular}\label{Table-Self}
\end{table}

\subsubsection{The Effectiveness of Auxiliary Self-Supervision}
To demonstrate the necessity and superiority of the auxiliary self-supervision signals, we report the performance of our DTGRM model and its variants with or without auxiliary self-supervision signal during training stage on 50Salads dataset. As we can see in Table \ref{Table-Self}, the models trained with self-supervision signals outperforms their duplicates, which are trained only with ground-truth action segmentation labels. Specifically, the ``S-Graph'' performs very bad because the constructed S-Graph usually is very noisy, while its performance is improved by a large margin after trained with self-supervision signal. This shows that the proposed self-supervised task can improve the model's generalization ability and make it more robust to the noise in the constructed graphs. Moreover, besides our DTGRM, the self-supervision task is also helpful for temporal convolution (DTCN), which indicates that the self-supervision task can be widely used to other action segmentation models. From the qualitative comparison in Fig. \ref{Fig-compare-self}, we can see that the auxiliary self-supervised task is very useful to improve the quality of action segmentation results. Especially, the model trained with self-supervision is able to correct the wrong action segments with considerable duration and reduce the over-segmentation errors at the boundaries of action segments. 

\subsubsection{The Impact of Hyper-Parameters}
The effect of the proposed auxiliary self-supervised task is controlled by three hyper-parameters: $\lambda_c$, $\lambda_e$ and $\eta$. As shown in Table \ref{Table-impact-paras}, in this section, we study the impact of these parameters and see how they affect the performance of the proposed model. In all experiments in Table \ref{Table-impact-paras}, we set $\omega = 0.15$, whose impact has been fully analyzed in MSTCN.

To analyze the effect of the parameter $\lambda_c$, we train different models with different values of $\lambda_c$ and $\lambda_e = 2$. As we can see in Table \ref{Table-impact-paras}, the impact of $\lambda_c$ is relatively small on the performance when $\lambda_c > 0$. Compared the model without self-supervision, i.e., DTGRM(w/o self) in Table \ref{Table-Self}, increasing $\lambda_c$ to 1.0 still improves the performance but not as good as the value of $\lambda_c=0.5$. However, there is a huge degradation in performance when we reduce $\lambda_c$ to 0, which indicates that the $\mathcal{L}_{corr}$ is an essential factor for action segmentation task. The hyper-parameter $\lambda_e$ is another parameter that balance the multiple components in the loss function of the self-supervised task. Our default value is $\lambda_e = 2$. While the values $\lambda_e = 1, 3$ still gives an improvement over the baseline without self-supervision, setting $\lambda_e=4$ results in a huge drop in performance. This may be because large $\lambda_e$ makes the loss function focus on finding the exchanged frames while ignoring correcting them.

In addition, the hyper-parameter $\eta$ defines the number of the exchanged frames in video, which play a vital role in auxiliary self-supervised task. As shown in Table \ref{Table-impact-paras}, increasing $\eta$ from 5 to 20 significantly improves the performance. This is mainly because the exchanged frames perfectly simulate the over-segmentation errors and make the model explicitly penalizes them. However, when there are too many exchanged frames (i.e., $\eta = 30$), the model performs worse since the correct temporal relations in video have been disturbed heavily. More ablation studies on hyperparameter number of level $K$ and the order of graph $O_t$ are included in the supplementary materials.

\begin{table}[htbp]
	\centering
	\caption{Impact of hyper-parameters $\lambda_c$, $\lambda_e$ and $\eta$ on the 50Salads dataset. More results are in the \emph{Supp. Materials}.}
	\vspace{-0.2cm}
	\begin{tabular}{cccccc}
		\hline
		\textbf{Impact of $\lambda_c$} &\multicolumn{3}{c}{\textbf{F1}$@\{ \text{10,25,}50 \}$} &\textbf{Edit} &\textbf{Acc} \\  
		\hline
		($\lambda_c=0.0$, $\lambda_e=2$) &58.7 &55.3 &46.0 &53.0 &72.1 \\
		($\lambda_c=0.5$, $\lambda_e=2$) &\textbf{79.1} &\textbf{75.9} &\textbf{66.1} &\textbf{72.0} &\textbf{80.0} \\
		($\lambda_c=1.0$, $\lambda_e=2$) &77.7 &74.7 &64.4 &71.4 &78.3 \\
		\hline
		\hline
		\textbf{Impact of $\lambda_e$} &\multicolumn{3}{c}{\textbf{F1}$@\{ \text{10,25,}50 \}$} &\textbf{Edit} &\textbf{Acc} \\ 
		\hline
		($\lambda_c=0.5$, $\lambda_e=1$) &77.2 &74.7 &64.6 &70.4 &78.7 \\
		($\lambda_c=0.5$, $\lambda_e=2$) &\textbf{79.1} &\textbf{75.9} &\textbf{66.1} &\textbf{72.0} &\textbf{80.0} \\
		($\lambda_c=0.5$, $\lambda_e=3$) &74.3 &71.4 &62.5 &66.2 &78.6 \\
		\hline
		\hline
		\textbf{Impact of $\eta$} &\multicolumn{3}{c}{\textbf{F1}$@\{ \text{10,25,}50 \}$} &\textbf{Edit} &\textbf{Acc} \\ 
		\hline
		$\eta=10$ &76.6 &73.0 &63.2 &69.7 &78.5 \\
		$\eta=20$ &\textbf{79.1} &\textbf{75.9} &\textbf{66.1} &\textbf{72.0} &\textbf{80.0} \\
		$\eta=30$ &77.2 &74.6 &65.7 &71.1 &79.2 \\
		\hline
	\end{tabular}\label{Table-impact-paras}
\end{table}

\section{Conclusion}
In this paper, we propose to model the short and long-range temporal relations in action segmentation. We construct multi-level dilated temporal graphs to capture the temporal relations in various time spans and propose DRGC layers to perform relational reasoning. Further, an auxiliary self-supervision is introduced to explicitly simulate the over-segmentation errors in predictions. Extensive experiments showed that our model can effectively conduct temporal relational reasoning in different timescales, and outperform the state-of-the-art methods on three challenging datasets.


\section*{Acknowledgement}
This work was supported in part by the Beijing Outstanding Young Scientist Program NO. BJJWZYJH012019100020098, the Fundamental Research Funds for the Central Universities, the Research Funds of Renmin University of China, and Public Computing Cloud, Renmin University of China.

\bibliographystyle{aaai21}
\bibliography{refs}

\begin{thebibliography}{46}
\providecommand{\natexlab}[1]{#1}
\providecommand{\url}[1]{\texttt{#1}}
\providecommand{\urlprefix}{URL }
\expandafter\ifx\csname urlstyle\endcsname\relax
  \providecommand{\doi}[1]{doi:\discretionary{}{}{}#1}\else
  \providecommand{\doi}{doi:\discretionary{}{}{}\begingroup
  \urlstyle{rm}\Url}\fi

\bibitem[{Carreira and Zisserman(2017)}]{carreira2017quo}
Carreira, J.; and Zisserman, A. 2017.
\newblock Quo vadis, action recognition? a new model and the kinetics dataset.
\newblock In \emph{IEEE Conference on Computer Vision and Pattern Recognition},
  6299--6308.

\bibitem[{Chen et~al.(2020)Chen, Li, Bao, and AlRegib}]{chen2020action}
Chen, M.-H.; Li, B.; Bao, Y.; and AlRegib, G. 2020.
\newblock Action Segmentation with Mixed Temporal Domain Adaptation.
\newblock In \emph{IEEE Winter Conference on Applications of Computer Vision},
  605--614.

\bibitem[{Chen et~al.(2019)Chen, Rohrbach, Yan, Shuicheng, Feng, and
  Kalantidis}]{chen2019graph}
Chen, Y.; Rohrbach, M.; Yan, Z.; Shuicheng, Y.; Feng, J.; and Kalantidis, Y.
  2019.
\newblock Graph-based global reasoning networks.
\newblock In \emph{IEEE Conference on Computer Vision and Pattern Recognition},
  433--442.

\bibitem[{Cheng et~al.(2014)Cheng, Fan, Pankanti, and
  Choudhary}]{cheng2014temporal}
Cheng, Y.; Fan, Q.; Pankanti, S.; and Choudhary, A. 2014.
\newblock Temporal sequence modeling for video event detection.
\newblock In \emph{IEEE Conference on Computer Vision and Pattern Recognition},
  2227--2234.

\bibitem[{Danafar and Gheissari(2007)}]{danafar2007action}
Danafar, S.; and Gheissari, N. 2007.
\newblock Action recognition for surveillance applications using optic flow and
  SVM.
\newblock In \emph{Asian Conference on Computer Vision}, 457--466.

\bibitem[{Doersch, Gupta, and Efros(2015)}]{doersch2015unsupervised}
Doersch, C.; Gupta, A.; and Efros, A.~A. 2015.
\newblock Unsupervised visual representation learning by context prediction.
\newblock In \emph{IEEE International Conference on Computer Vision},
  1422--1430.

\bibitem[{Farha and Gall(2019)}]{farha2019ms}
Farha, Y.~A.; and Gall, J. 2019.
\newblock Ms-tcn: Multi-stage temporal convolutional network for action
  segmentation.
\newblock In \emph{IEEE Conference on Computer Vision and Pattern Recognition},
  3575--3584.

\bibitem[{Fathi, Farhadi, and Rehg(2011)}]{fathi2011understanding}
Fathi, A.; Farhadi, A.; and Rehg, J.~M. 2011.
\newblock Understanding egocentric activities.
\newblock In \emph{International Conference on Computer Vision}, 407--414.

\bibitem[{Fathi and Rehg(2013)}]{fathi2013modeling}
Fathi, A.; and Rehg, J.~M. 2013.
\newblock Modeling actions through state changes.
\newblock In \emph{IEEE Conference on Computer Vision and Pattern Recognition},
  2579--2586.

\bibitem[{Fathi, Ren, and Rehg(2011)}]{fathi2011learning}
Fathi, A.; Ren, X.; and Rehg, J.~M. 2011.
\newblock Learning to recognize objects in egocentric activities.
\newblock In \emph{IEEE Conference on Computer Vision and Pattern Recognition},
  3281--3288.

\bibitem[{Fernando et~al.(2017)Fernando, Bilen, Gavves, and
  Gould}]{fernando2017self}
Fernando, B.; Bilen, H.; Gavves, E.; and Gould, S. 2017.
\newblock Self-supervised video representation learning with odd-one-out
  networks.
\newblock In \emph{IEEE Conference on Computer Vision and Pattern Recognition},
  3636--3645.

\bibitem[{Gidaris, Singh, and Komodakis(2018)}]{gidaris2018unsupervised}
Gidaris, S.; Singh, P.; and Komodakis, N. 2018.
\newblock Unsupervised representation learning by predicting image rotations.
\newblock In \emph{International Conference on Learning Representations}.

\bibitem[{Hu, Nie, and Li(2019)}]{hu2019deep}
Hu, D.; Nie, F.; and Li, X. 2019.
\newblock Deep multimodal clustering for unsupervised audiovisual learning.
\newblock In \emph{Proceedings of the IEEE Conference on Computer Vision and
  Pattern Recognition}, 9248--9257.

\bibitem[{Hu et~al.(2020)Hu, Qian, Jiang, Tan, Wen, Ding, Lin, and
  Dou}]{hu2020discriminative}
Hu, D.; Qian, R.; Jiang, M.; Tan, X.; Wen, S.; Ding, E.; Lin, W.; and Dou, D.
  2020.
\newblock Discriminative Sounding Objects Localization via Self-supervised
  Audiovisual Matching.
\newblock \emph{Advances in Neural Information Processing Systems} 33.

\bibitem[{Huang, Fei-Fei, and Niebles(2016)}]{huang2016connectionist}
Huang, D.-A.; Fei-Fei, L.; and Niebles, J.~C. 2016.
\newblock Connectionist temporal modeling for weakly supervised action
  labeling.
\newblock In \emph{European Conference on Computer Vision}, 137--153.

\bibitem[{Huang, Sugano, and Sato(2020)}]{huang2020improving}
Huang, Y.; Sugano, Y.; and Sato, Y. 2020.
\newblock Improving Action Segmentation via Graph-Based Temporal Reasoning.
\newblock In \emph{IEEE Conference on Computer Vision and Pattern Recognition},
  14024--14034.

\bibitem[{Hussein, Gavves, and Smeulders(2019)}]{hussein2019videograph}
Hussein, N.; Gavves, E.; and Smeulders, A.~W. 2019.
\newblock Videograph: Recognizing minutes-long human activities in videos.
\newblock \emph{arXiv preprint arXiv:1905.05143} .

\bibitem[{Karaman, Seidenari, and Del~Bimbo(2014)}]{karaman2014fast}
Karaman, S.; Seidenari, L.; and Del~Bimbo, A. 2014.
\newblock Fast saliency based pooling of fisher encoded dense trajectories.
\newblock \emph{European Conference on Computer Vision THUMOS Workshop} 1(2):
  5.

\bibitem[{Kipf and Welling(2017)}]{kipf2016semi}
Kipf, T.~N.; and Welling, M. 2017.
\newblock Semi-supervised classification with graph convolutional networks.
\newblock In \emph{International Conference on Learning Representations}.

\bibitem[{Koppula and Saxena(2015)}]{koppula2015anticipating}
Koppula, H.~S.; and Saxena, A. 2015.
\newblock Anticipating human activities using object affordances for reactive
  robotic response.
\newblock \emph{IEEE Transactions on Pattern Analysis and Machine Intelligence}
  38(1): 14--29.

\bibitem[{Korbar, Tran, and Torresani(2018)}]{korbar2018cooperative}
Korbar, B.; Tran, D.; and Torresani, L. 2018.
\newblock Cooperative learning of audio and video models from self-supervised
  synchronization.
\newblock In \emph{Advances in Neural Information Processing Systems},
  7763--7774.

\bibitem[{Kr{\"u}ger et~al.(2007)Kr{\"u}ger, Kragic, Ude, and
  Geib}]{kruger2007meaning}
Kr{\"u}ger, V.; Kragic, D.; Ude, A.; and Geib, C. 2007.
\newblock The meaning of action: A review on action recognition and mapping.
\newblock \emph{Advanced Robotics} 21(13): 1473--1501.

\bibitem[{Kuehne, Arslan, and Serre(2014)}]{kuehne2014language}
Kuehne, H.; Arslan, A.; and Serre, T. 2014.
\newblock The language of actions: Recovering the syntax and semantics of
  goal-directed human activities.
\newblock In \emph{IEEE Conference on Computer Vision and Pattern Recognition},
  780--787.

\bibitem[{Kuehne, Gall, and Serre(2016)}]{kuehne2016end}
Kuehne, H.; Gall, J.; and Serre, T. 2016.
\newblock An end-to-end generative framework for video segmentation and
  recognition.
\newblock In \emph{IEEE Winter Conference on Applications of Computer Vision},
  1--8.

\bibitem[{Lea et~al.(2017)Lea, Flynn, Vidal, Reiter, and
  Hager}]{lea2017temporal}
Lea, C.; Flynn, M.~D.; Vidal, R.; Reiter, A.; and Hager, G.~D. 2017.
\newblock Temporal convolutional networks for action segmentation and
  detection.
\newblock In \emph{IEEE Conference on Computer Vision and Pattern Recognition},
  156--165.

\bibitem[{Lea et~al.(2016)Lea, Reiter, Vidal, and Hager}]{lea2016segmental}
Lea, C.; Reiter, A.; Vidal, R.; and Hager, G.~D. 2016.
\newblock Segmental spatiotemporal cnns for fine-grained action segmentation.
\newblock In \emph{European Conference on Computer Vision}, 36--52.

\bibitem[{Lea, Vidal, and Hager(2016)}]{lea2016learning}
Lea, C.; Vidal, R.; and Hager, G.~D. 2016.
\newblock Learning convolutional action primitives for fine-grained action
  recognition.
\newblock In \emph{IEEE International Conference on Robotics and Automation},
  1642--1649.

\bibitem[{Lee et~al.(2017)Lee, Huang, Singh, and Yang}]{lee2017unsupervised}
Lee, H.-Y.; Huang, J.-B.; Singh, M.; and Yang, M.-H. 2017.
\newblock Unsupervised representation learning by sorting sequences.
\newblock In \emph{IEEE International Conference on Computer Vision}, 667--676.

\bibitem[{Lei and Todorovic(2018)}]{lei2018temporal}
Lei, P.; and Todorovic, S. 2018.
\newblock Temporal deformable residual networks for action segmentation in
  videos.
\newblock In \emph{IEEE Conference on Computer Vision and Pattern Recognition},
  6742--6751.

\bibitem[{Li et~al.(2020)Li, AbuFarha, Liu, Cheng, and Gall}]{li2020ms}
Li, S.-J.; AbuFarha, Y.; Liu, Y.; Cheng, M.-M.; and Gall, J. 2020.
\newblock MS-TCN++: Multi-Stage Temporal Convolutional Network for Action
  Segmentation.
\newblock \emph{IEEE Transactions on Pattern Analysis and Machine Intelligence}
  \doi{10.1109/TPAMI.2020.3021756}.

\bibitem[{Li and Gupta(2018)}]{li2018beyond}
Li, Y.; and Gupta, A. 2018.
\newblock Beyond grids: Learning graph representations for visual recognition.
\newblock In \emph{Advances in Neural Information Processing Systems},
  9225--9235.

\bibitem[{Liang et~al.(2018)Liang, Hu, Zhang, Lin, and
  Xing}]{liang2018symbolic}
Liang, X.; Hu, Z.; Zhang, H.; Lin, L.; and Xing, E.~P. 2018.
\newblock Symbolic graph reasoning meets convolutions.
\newblock In \emph{Advances in Neural Information Processing Systems},
  1853--1863.

\bibitem[{Misra, Zitnick, and Hebert(2016)}]{misra2016shuffle}
Misra, I.; Zitnick, C.~L.; and Hebert, M. 2016.
\newblock Shuffle and learn: unsupervised learning using temporal order
  verification.
\newblock In \emph{European Conference on Computer Vision}, 527--544.

\bibitem[{Pirsiavash and Ramanan(2014)}]{pirsiavash2014parsing}
Pirsiavash, H.; and Ramanan, D. 2014.
\newblock Parsing videos of actions with segmental grammars.
\newblock In \emph{IEEE Conference on Computer Vision and Pattern Recognition},
  612--619.

\bibitem[{Rasouli and Tsotsos(2019)}]{rasouli2019autonomous}
Rasouli, A.; and Tsotsos, J.~K. 2019.
\newblock Autonomous vehicles that interact with pedestrians: A survey of
  theory and practice.
\newblock \emph{IEEE Transactions on Intelligent Transportation Systems} 21(3):
  900--918.

\bibitem[{Rohrbach et~al.(2012)Rohrbach, Amin, Andriluka, and
  Schiele}]{rohrbach2012database}
Rohrbach, M.; Amin, S.; Andriluka, M.; and Schiele, B. 2012.
\newblock A database for fine grained activity detection of cooking activities.
\newblock In \emph{IEEE Conference on Computer Vision and Pattern Recognition},
  1194--1201.

\bibitem[{Sadigh et~al.(2016)Sadigh, Sastry, Seshia, and
  Dragan}]{sadigh2016planning}
Sadigh, D.; Sastry, S.; Seshia, S.~A.; and Dragan, A.~D. 2016.
\newblock Planning for autonomous cars that leverage effects on human actions.
\newblock In \emph{Robotics: Science and Systems}, volume~2.

\bibitem[{Shen et~al.(2018)Shen, Li, Yi, Chen, and Wang}]{shen2018person}
Shen, Y.; Li, H.; Yi, S.; Chen, D.; and Wang, X. 2018.
\newblock Person re-identification with deep similarity-guided graph neural
  network.
\newblock In \emph{European Conference on Computer Vision}, 486--504.

\bibitem[{Singh et~al.(2016)Singh, Marks, Jones, Tuzel, and
  Shao}]{singh2016multi}
Singh, B.; Marks, T.~K.; Jones, M.; Tuzel, O.; and Shao, M. 2016.
\newblock A multi-stream bi-directional recurrent neural network for
  fine-grained action detection.
\newblock In \emph{IEEE Conference on Computer Vision and Pattern Recognition},
  1961--1970.

\bibitem[{Stein and McKenna(2013)}]{stein2013combining}
Stein, S.; and McKenna, S.~J. 2013.
\newblock Combining embedded accelerometers with computer vision for
  recognizing food preparation activities.
\newblock In \emph{ACM International Joint Conference on Pervasive and
  Ubiquitous Computing}, 729--738.

\bibitem[{Wang and Gupta(2018)}]{wang2018videos}
Wang, X.; and Gupta, A. 2018.
\newblock Videos as space-time region graphs.
\newblock In \emph{European conference on computer vision}, 399--417.

\bibitem[{Wang et~al.(2020)Wang, Gao, Wang, Li, and Wu}]{wangboundary}
Wang, Z.; Gao, Z.; Wang, L.; Li, Z.; and Wu, G. 2020.
\newblock Boundary-Aware Cascade Networks for Temporal Action Segmentation.
\newblock In \emph{European Conference on Computer Vision}, 36--52.

\bibitem[{Yan, Xiong, and Lin(2018)}]{yan2018spatial}
Yan, S.; Xiong, Y.; and Lin, D. 2018.
\newblock Spatial temporal graph convolutional networks for skeleton-based
  action recognition.
\newblock \emph{arXiv preprint arXiv:1801.07455} .

\bibitem[{Zeng et~al.(2019)Zeng, Huang, Tan, Rong, Zhao, Huang, and
  Gan}]{zeng2019graph}
Zeng, R.; Huang, W.; Tan, M.; Rong, Y.; Zhao, P.; Huang, J.; and Gan, C. 2019.
\newblock Graph convolutional networks for temporal action localization.
\newblock In \emph{IEEE International Conference on Computer Vision},
  7094--7103.

\bibitem[{Zhang et~al.(2020)Zhang, Shen, Xu, and Shen}]{zhang2020temporal}
Zhang, J.; Shen, F.; Xu, X.; and Shen, H.~T. 2020.
\newblock Temporal reasoning graph for activity recognition.
\newblock \emph{IEEE Transactions on Image Processing} 29: 5491--5506.

\bibitem[{Zhang et~al.(2019)Zhang, Tokmakov, Hebert, and
  Schmid}]{zhang2019structured}
Zhang, Y.; Tokmakov, P.; Hebert, M.; and Schmid, C. 2019.
\newblock A structured model for action detection.
\newblock In \emph{IEEE Conference on Computer Vision and Pattern Recognition},
  9975--9984.

\end{thebibliography}

\end{document}